\documentclass[sigconf]{acmart}

\AtBeginDocument{%
  \providecommand\BibTeX{{%
    \normalfont B\kern-0.5em{\scshape i\kern-0.25em b}\kern-0.8em\TeX}}}

\setcopyright{none}

\acmConference[HIP'23]{7th International Workshop on Historical Document Imaging and Processing}{August 25--26,
  2023}{San José, California}
\usepackage{multirow}
\usepackage{multicol}
\usepackage{subcaption,graphicx}
\usepackage{chngcntr}
\usepackage{acronym}

\acrodef{CNN}{Convolutional Neural Network}
\acrodef{OCR}{Optical Character Recognition}
\acrodef{WPI}{Wildenstein Plattner Institute}

\begin{document}

\title{DocLangID: Improving Few-Shot Training to Identify the Language of Historical Documents}


\author{Furkan Simsek}
\email{furkan.simsek@student.hpi.de}
\affiliation{%
  \institution{Hasso Plattner Institute}
  \city{Potsdam}
  \country{Germany}
}

\author{Brian Pfitzmann}
\email{brian.pfitzmann@student.hpi.de}
\affiliation{%
  \institution{Hasso Plattner Institute}
  \city{Potsdam}
  \country{Germany}
}

\author{Hendrik Raetz}
\email{hendrik.raetz@hpi.de}
\affiliation{%
  \institution{Hasso Plattner Institute}
  \city{Potsdam}
  \country{Germany}
}

\author{Jona Otholt}
\email{jona.otholt@hpi.de}
\affiliation{%
  \institution{Hasso Plattner Institute}
  \city{Potsdam}
  \country{Germany}
}

\author{Haojin Yang}
\email{haojin.yang@hpi.de}
\affiliation{%
  \institution{Hasso Plattner Institute}
  \city{Potsdam}
  \country{Germany}
}

\author{Christoph Meinel}
\email{christoph.meinel@hpi.de}
\affiliation{%
  \institution{Hasso Plattner Institute}
  \city{Potsdam}
  \country{Germany}
}

\renewcommand{\shortauthors}{Simsek and Pfitzmann, et al.}

\begin{abstract}
Language identification describes the task of recognizing the language of written text in documents. This information is crucial because it can be used to support the analysis of a document's vocabulary and context. 
Supervised learning methods in recent years have advanced the task of language identification. 
However, these methods usually require large labeled datasets, which often need to be included for various domains of images, such as documents or scene images. 
In this work, we propose \textit{DocLangID}, a transfer learning approach to identify the language of unlabeled historical documents. 
We achieve this by first leveraging labeled data from a different but related domain of historical documents. 
Secondly, we implement a distance-based few-shot learning approach to adapt a convolutional neural network to new languages of the unlabeled dataset. 
By introducing small amounts of manually labeled examples from the set of unlabeled images, our feature extractor develops a better adaptability towards new and different data distributions of historical documents. 
We show that such a model can be effectively fine-tuned for the unlabeled set of images by only reusing the same few-shot examples. 
We showcase our work across 10 languages that mostly use the latin script. 
Our experiments on historical documents demonstrate that our combined approach improves the language identification performance, achieving 74\% recognition accuracy on the four unseen languages of the unlabeled dataset.
\end{abstract}




\begin{teaserfigure}
  \centering
  \includegraphics[width=0.9\textwidth, height=250pt]{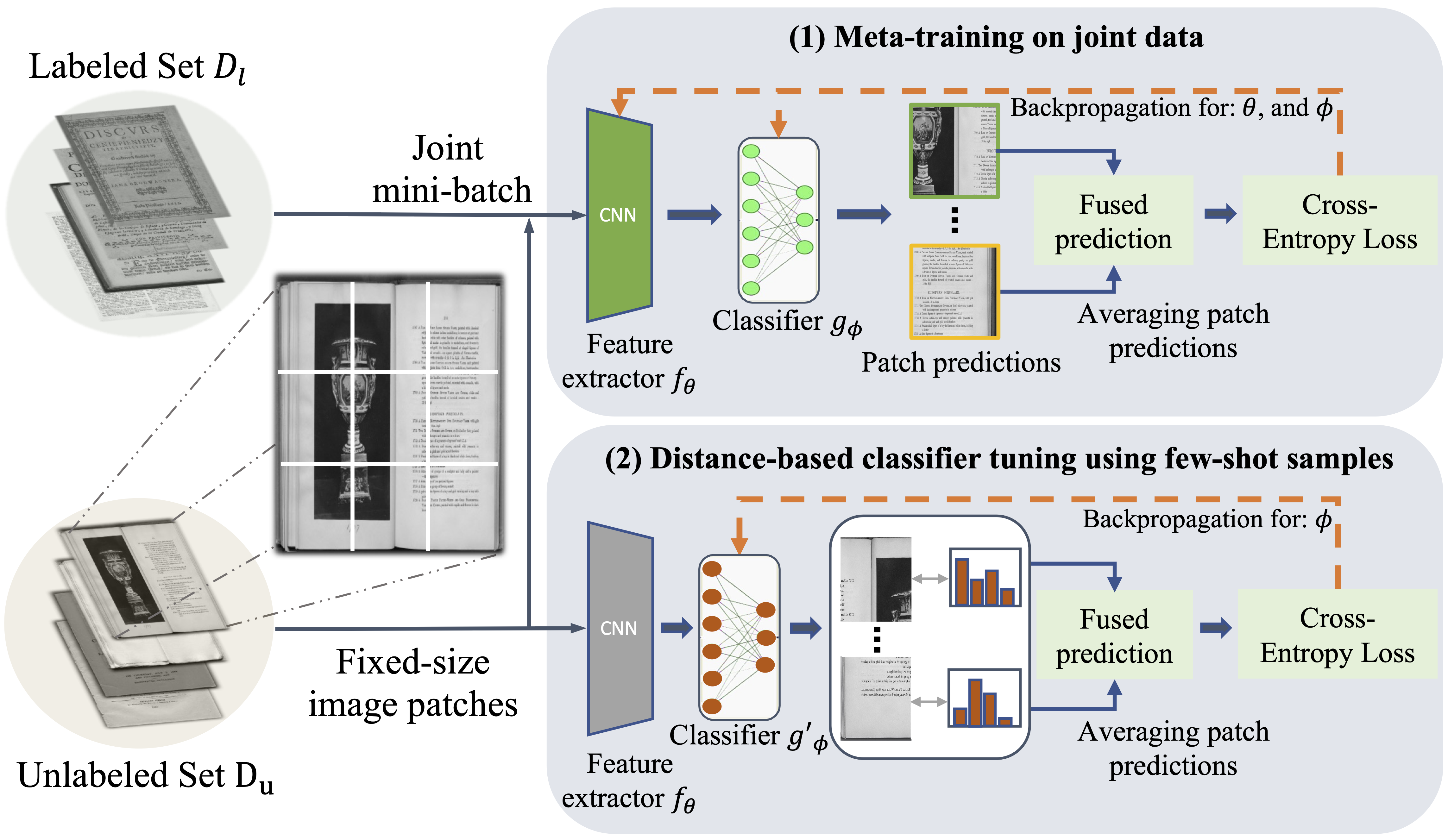}
  \caption{Overview of \textit{DocLangID}: This two-stage approach starts with training a CNN on a joint dataset containing numerous labeled samples from a source domain dataset $D_l$ and a few manually labeled samples from an unlabeled target domain dataset $D_u$. In the second stage, we fix the feature extractor parameters and replace the original classifier with a new distance-based classifier. This new classifier is fine-tuned on the same few-shot samples from the previous stage, leveraging the source domain knowledge of the feature extractor to help adapt to the target domain.}
  \label{fig:teaser}
\end{teaserfigure}


\maketitle

\section{Introduction}
Language identification is a subfield of image classification and aims to recognize the language of printed and written text appearing in images. One exciting and less researched type of image is scans of historical documents. For instance, a significant and diverse collection of text-based historical documents compiled by several major European libraries can be accessed in the IMPACT \cite{impact_paper} dataset. The creation of such a dataset, together with accompanying ground-truth data for a smaller subset, is often connected to the goal of improving the analysis and digital processing of historical documents. Although many historical documents primarily contain text, such as text from books, articles, or newsletters, there are also documents whose main content is represented through large pictures. One specific example is art-historical documents. These documents are written records that provide information about works of art and the artists who created them. They include manuscripts, letters, diaries, and other written materials that provide insights into a particular artwork's creation, history, and significance.

Being one of the broader discussed Computer Vision tasks, \ac{OCR} methods aim to convert the text in images into a digital format. The language of a document is a valuable and contextual information for the process of character recognition. For example, if the language of the document is known in prior, a language-specific \ac{OCR} engine can be selected. This can lead to reducing the amount of ambiguity in character recognition, likely improving recognition quality and speed. 

In previous years, deep learning methods have achieved significant performance improvements on various document analysis problems, such as image classification and \ac{OCR} \cite{super_resoulution, Yin2017SceneTR, Lyu2018MaskTA}. However, there are multiple aspects of document input images that still weaken the recognition performance of the above methods. A poor image resolution, small font sizes, and noise introduced by the input image itself or due to the scanning process are common problems \cite{super_resoulution, Yin2017SceneTR, Lyu2018MaskTA} and, at the same time, common characteristics of art-historical documents. Several supervised learning methods were proposed to tackle these problems and improve recognition performance \cite{Yin2017SceneTR, Lyu2018MaskTA}. These methods usually incorporate large amounts of labeled data into the training of a neural network. However, most of today's data is generated from diverse, unstructured sources, resulting in data without additional ground truth information. Therefore, accurately assigning meaningful labels often requires contextual understanding and human expertise, which is usually time-consuming and infeasible for large quantities of data.

Motivated by this problem, in this paper, we propose \textit{DocLangID}, a two-stage training approach to tackle the problem of domain adaptation for language identification. Given two domains of images, one labeled and the other unlabeled, the task is to adapt a model trained on the labeled domain toward identifying the languages of the unlabeled domain. In this setting, the languages of both domains do not have to overlap. 

\textit{DocLangID} is a simple yet effective few-shot learning method that employs a distance-based classifier, as shown by Chen et al. \cite{closerlook}, to improve language identification in the unlabeled domain. We achieve this improvement by using only small amounts of manually labeled examples.

Our contributions can be summarized as follows:

\begin{enumerate}
    \item We achieve high language identification performance for art-historical documents in a supervised setting, which was not accomplished previously for languages of the Latin script.
    \item We develop a simple but effective method for domain adaptation, which utilizes large amounts of labeled data and few-shot examples to transfer the model's knowledge to a new domain of historical documents.
    \item We analyze patch extraction, a commonly used procedure when training on image data, and the assessment of its impact on our approach.
    \item We implement and evaluate supervised variations of our main \textit{DocLangID} approach, and we publish our work on Github \footnote{\url{https://github.com/caesarea38/DocLangID}}.
\end{enumerate}
\section{Related work}
\textbf{Language Identification} Several machine learning-based methods in previous years have tackled the problem of language identification from images \cite{diacritics_on_device,lidensemblecnn,hierarchicalclassification}. Vatsal et al. \cite{diacritics_on_device} developed a \ac{CNN} used to identify the language based on the presence of diacritic characters in the image. Shah et al. \cite{hierarchicalclassification} use multiple \ac{CNN}s in a hierarchy to identify six Indian languages belonging to two different language groups. They use a first \ac{CNN} as a binary classification model to identify the correct language group, followed by another \ac{CNN} used to identify the exact language. Moreover, an ensemble approach was proposed by Chakraborty et al. \cite{lidensemblecnn} that, based on the RGB color model, generates five different samples from a scene text and combines the predictions for each sample to improve recognition performance. Furthermore, identifying the script used in images is another critical related area, often considered an essential preliminary step for text understanding systems \cite{localcnnforscriptid,recentscriptidpaper}. A combination of a local \ac{CNN} for exploiting local image features and a global \ac{CNN} for general image features was introduced by Lu et al. \cite{localcnnforscriptid}. More recently, Mahajan et al. \cite{recentscriptidpaper} employed \ac{CNN}s for a word-level script identification approach and applied it to scenic images with Indian script texts.

\textbf{Few-Shot Learning} Few-shot learning is a type of machine learning which involves training a model using a tiny number of (labeled) examples and aims to improve model generalization on unseen data. To overcome the costly labeling process of large unlabeled datasets, various types of few-shot learning algorithms for neural networks were proposed that we briefly review next. Finn et al. \cite{Finn2017ModelAgnosticMF} proposed a meta-learning approach that aims to train a model on various tasks, such that quick adaptation to a new task can be achieved using small amounts of labeled examples and a small number of gradient update steps. Additionally, Rusu et al. \cite{Rusu2018MetaLearningWL} propose a latent embedding optimization method that performs gradient-based meta-learning on the embedding space of the model parameters. Furthermore, distance metric learning-based methods \cite{qilowshotlearning,gidaris} address the few-shot learning problem by "learning to compare". The models are trained to classify an unseen input image by determining its similarity to the few labeled examples during training. Common examples of distance metrics used during training are cosine similarity \cite{cosinesimilarity} and euclidean distance to class-mean representation \cite{euclideandistancetoclassmeanrep}. However, previous methods were typically applied and evaluated on standard benchmarks like CIFAR \cite{Krizhevsky2009LearningML}, neglecting realistic constraints and challenges in historical documents, such as text comprehensibility affected by noise or color corruption. To the best of our knowledge, no prior work has yet explored the feasibility of language identification through supervised learning and domain adaptation, specifically within the context of art-historical documents containing languages from the Latin script.
\section{Method}
In this work, we investigate the problem of adapting a language identification model trained on a labeled source domain dataset ($D_l$) to a related but unlabeled target domain dataset ($D_u$). We frame this as a multiclass classification problem, with languages representing the classes. We consider label sets $Y_l$ for $D_l$ and $Y_u$ for $D_u$, which may share common languages.
\label{datasets}
\subsection{Datasets}
Throughout our work, we use two different datasets, which will be briefly presented in the following.

Our first dataset is a subset of the IMPACT \cite{impact_paper} dataset, which contains more than half a million images of text documents collected from major European libraries. 
The dataset includes various images with material from newspapers, books, pamphlets, and typewritten notes in multiple languages, including Dutch, French, Spanish, Czech, and Bulgarian, among others \cite{impact_paper}.

We created our second dataset from the publicly accessible online available of the \textit{\ac{WPI}} \footnote{\url{https://wpi.art}}, featuring over 21,000 scanned historical auction catalogs with one to two million unlabeled images from individual pages dating from the 17th to 20th century. We randomly selected a subset of these images, downloaded them, and manually labeled them by assigning the correct language. This resulted in a dataset containing 1,000 labeled images for each language (English, French, German, and Dutch), representing our target domain.

Our labeled IMPACT subset comprises exactly 3000 images, with 500 images per language. During training, we use 2100 images as training data (350 images per language) and the remaining 900 images as evaluation data. For our few-shot learning part, we manually selected 50 samples per \ac{WPI} language. The remaining \ac{WPI} data is used for evaluation.

\begin{figure}[!tbp]
  \centering
  \begin{subfigure}{0.5\columnwidth}
    \centering
    \includegraphics[width=\linewidth]{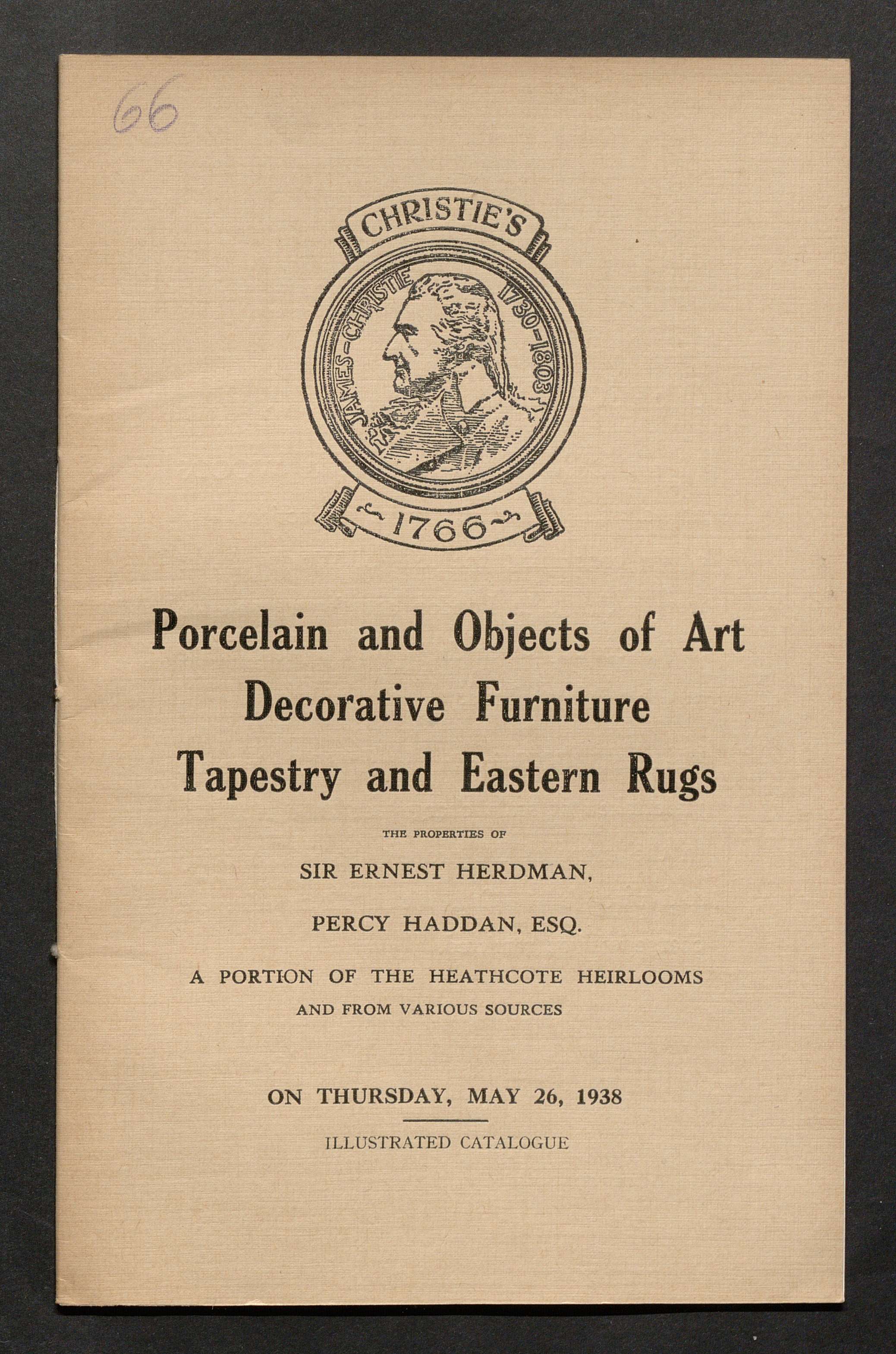}
    \caption{Raw image}
  \end{subfigure}%
  \hfill
  \begin{subfigure}{0.5\columnwidth}
    \centering
    \includegraphics[width=\linewidth]{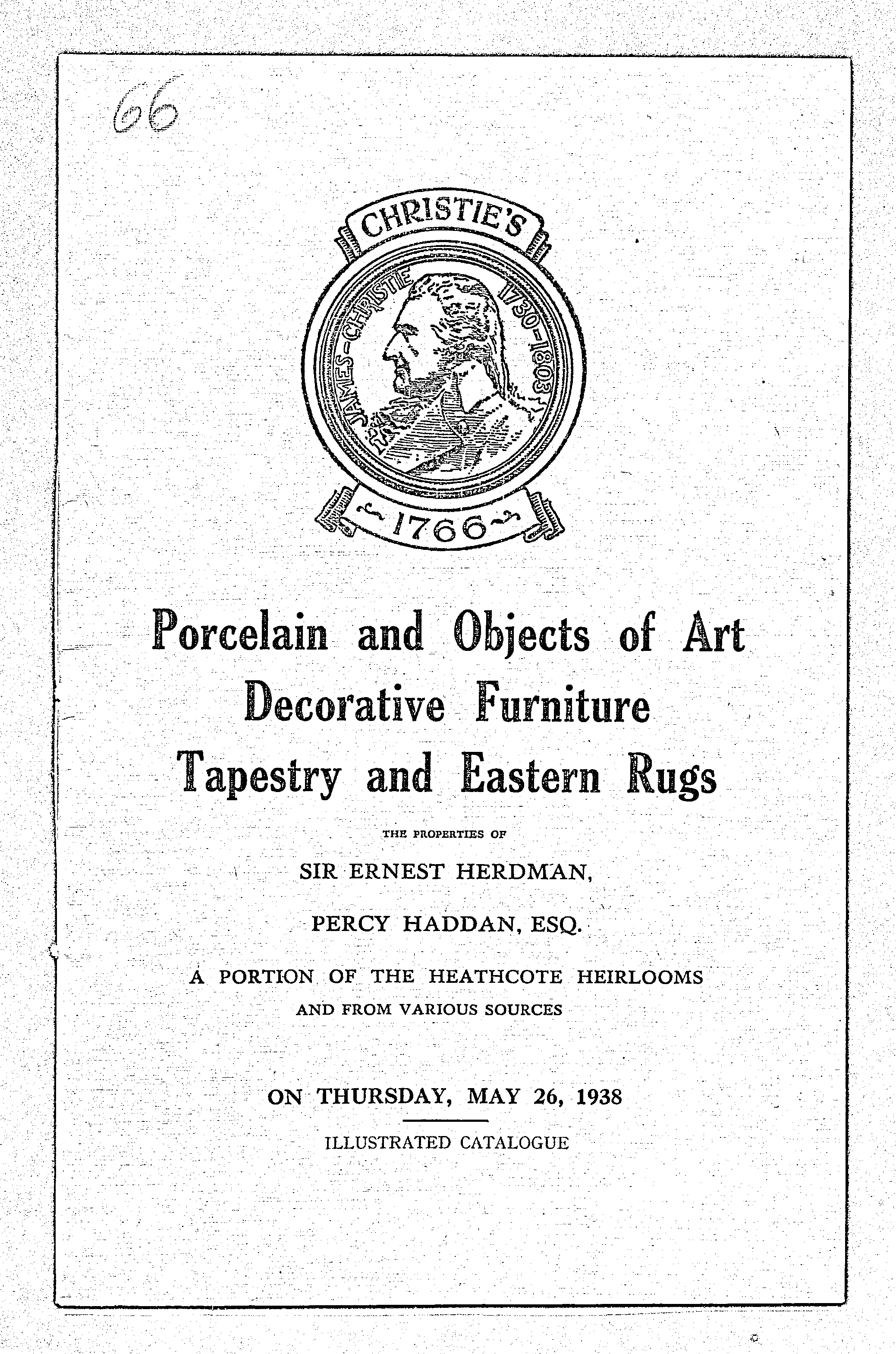}
    \caption{Binarized image
version}
  \end{subfigure}%
  \caption{Example image for the \ac{WPI} dataset without (a) preprocessing and after
preprocessing (b)}
  \label{fig:image_transformations}
\end{figure}

Prior to training our model, we apply two preprocessing steps on the original images for both IMPACT \cite{impact_paper} and \ac{WPI} datasets. Since old historical documents often contain noise and are also corrupted in color, these preprocessing steps aim to reduce noise and put the actual text more into the foreground of the image. Therefore, we first convert all images into a grayscale format and then apply adaptive binary thresholding to obtain binarized images in the end. An example output of this sequence of image transformations can be seen in \autoref{fig:image_transformations}.
\subsection{Feature learning}
Throughout our work, we use a ResNet-18 \ac{CNN} \cite{residual_learning} without pre-trained weights as our feature extractor. This decision is motivated firstly because ResNet-18 is a robust and widely known model for vision tasks and has demonstrated impressive performance on challenging image classification benchmarks such as ImageNet \cite{residual_learning} in the past years. Secondly, arguing from a practical perspective, ResNet-18 is a comparatively small neural network with around 11 million learnable parameters. Together with its architectural design and the smaller number of layers, ResNet-18 allows for faster training and inference.
\label{sec:patch-extraction}
\subsection{Patch extraction and prediction}
Given an input image during training, one crucial step of the prediction in \textit{DocLangID} is to divide that image into fixed-sized patches. In contrast to a single, large image, multiple smaller patches allow for more efficient parallel processing while maintaining the local spatial information through the fixed patch size, effectively enabling the equal contribution of each patch to the final prediction. We represent this patch size using the \textit{(height, width)} dimensions. For instance, during training, we use a size of (256, 256) per patch and an image size of (1024, 1024), which yields a maximum number of 16 patches. The number of patches to use during training can be controlled via a hyper-parameter. Starting from the top left part of the input image, we obtain patches by sequentially extracting \textit{(height, width)} many pixels. After extracting the patches, we calculate one forward pass for all patches. We then use an average over the probability distributions computed for each patch to obtain a fused probability distribution. Finally, we select the class with the highest probability in the fused distribution to obtain the predicted class for the complete image. The components and the full process for the computation of a prediction are visualized in \autoref{fig:teaser}.
\subsection{Our two-stage training approach}
Our main \textit{DocLangID} approach is separated into two training stages carried out sequentially. In the first meta-training stage, we train our \ac{CNN}s feature extractor $f_\theta$ (parametrized by its network parameters $\theta$) and a classification head $g_\phi$. This part is designed as a simple linear layer and projects the features to an output vector $\hat y = [c_1, c_2,...,c_k]$,  where $k$ equals the sum of the languages in $Y_{l}$ and $Y_{u}$, i.e., $k=|Y_{l}| + |Y_{u}|$. Then, we apply a softmax function on the output of the linear layer and use the standard cross-entropy loss function to obtain our final loss.

For the second stage of our approach, we first freeze the network parameters $\theta$, which is common practice in transfer learning. Secondly, we discard the linear classification head $g_\phi$ and retrain a new classifier $g'_\phi$. Also, we now use a distance-based design for the classifier by following \cite{closerlook,qilowshotlearning}. For an input image, we compute its cosine distance to the weight matrix of our classifier. We then optimize the parameters of the new classifier with respect to the cross-entropy loss applied to the output of the cosine distance and the labels of the few-shot samples. The intuition to use this design for our problem lies in the diverse image characteristics of our target domain dataset. Even though a set of images may belong to the same class, they might differ drastically in terms of appearance (presence of title pages, large pictures in the image, and different font sizes), leading to a potentially high and hence, problematic intra-class variance. Such a distance-based classifier design has the advantage of explicitly reducing intra-class variance by using the distance value to measure how close an image is to a particular class.

Considering the variations of \textit{DocLangID}, we use the term \textit{ResNet-FewShot} to refer to training the entire model (feature extractor + classifier) only on the \ac{WPI} few-shot samples. Also, we refer to \textit{ResNet-Meta} as the model evaluated after the first stage of our \textit{DocLangID} approach, in which only the joint dataset is used for training (see first training state in \autoref{fig:teaser}). The joint dataset refers to the combined dataset, which includes the labeled examples from the IMPACT subset and the few-shot samples from the \ac{WPI} subset. Considering this, our key insights are that even a few labeled examples are sufficient for a model to understand an unseen domain better and that the abundance of labeled examples from a related domain enhances this understanding during training.
\subsection{Tesseract \ac{OCR}}
As stated in the introduction, we employ the open-source Tesseract \ac{OCR} algorithm \cite{4376991} and the Python library fasttext-langdetect \cite{joulin2016bag, joulin2016fasttext} as a baseline for comparison with \textit{DocLangID}.

Tesseract can identify over 100 languages, support more than 30 scripts, and simultaneously recognize multiple languages, allowing multi-language input during execution. Using an \ac{OCR} algorithm may initially seem contradictory to improving \ac{OCR} through language identification. However, this approach yields impressive results on both IMPACT and \ac{WPI} datasets. Also, language identification does not require perfect text extraction, as correctly recognizing a majority of words is sufficient to determine the language.

We run Tesseract once or multiple times on the source image, using concatenated languages with the same script from our dataset as input. For the IMPACT dataset, for example, we could use languages from the Latin script in the first run and languages from the Cyrillic script in the second run. Running Tesseract multiple times ensures that at least one output has the correct input script, preventing incorrect language determination due to incorrect script usage.

We compare Tesseract's \ac{OCR} results' confidence values for each recognized word, expecting the highest confidence to occur when using the correct script's input languages. Consequently, we proceed with the Tesseract output with the highest confidence.

We employ the Python library fasttext-langdetect, using a Facebook-developed model trained on Wikipedia and other sources, to identify 176 languages from extracted text. 

However, this approach's main drawback is the lengthy identification process. Tesseract execution can take up to 30 seconds depending on text length, and multi-execution further extends this duration. Consequently, this method is not viable for large datasets with tens of thousands of images and primarily serves as a comparison baseline in terms of inference time.
\section{Experimental evaluation}
In the following, we demonstrate and discuss our results on the evaluation subsets of the IMPACT and \ac{WPI} datasets. For implementation details, we refer to the appendix.
\subsection{Recognition quality}
\begin{figure}
    \centering
    \includegraphics[scale=0.5]{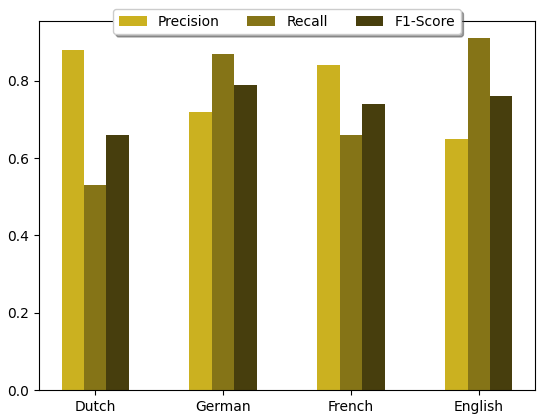}
    \caption{Classification metric results for each \ac{WPI} language (50 few-shot samples). Overall, despite minor differences between the languages, \textit{DocLangID} yields high classification metric results.}
    \label{fig:metrics_resnet}
\end{figure}

\begin{figure}[h!]
    \centering
    \includegraphics[width=\linewidth]{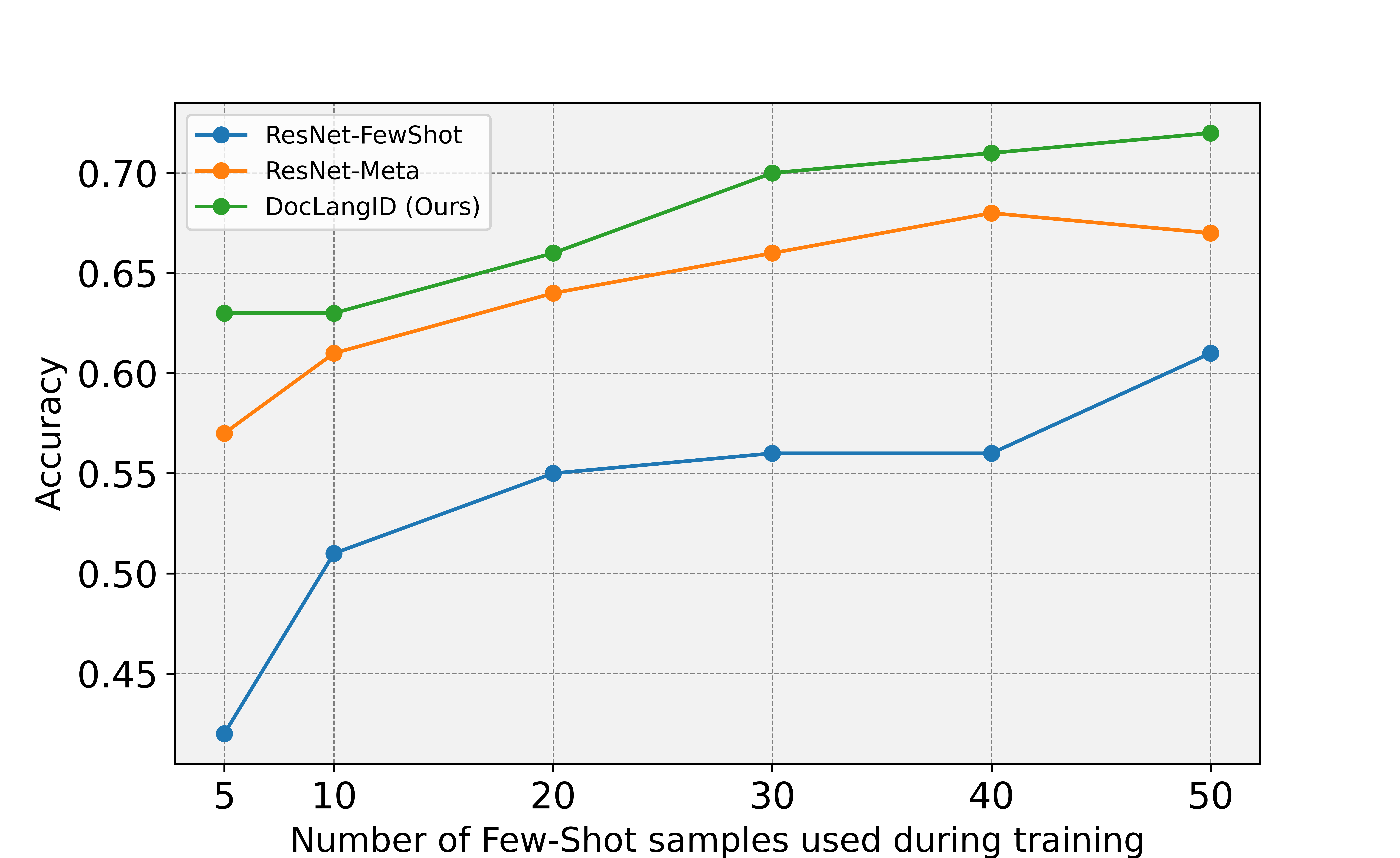}
    \caption{Influence of different numbers of few-shot samples on the recognition performance.}
    \label{fig:the-samples}
\end{figure}

\begin{figure}[]
    \centering
    \includegraphics[width=\linewidth]{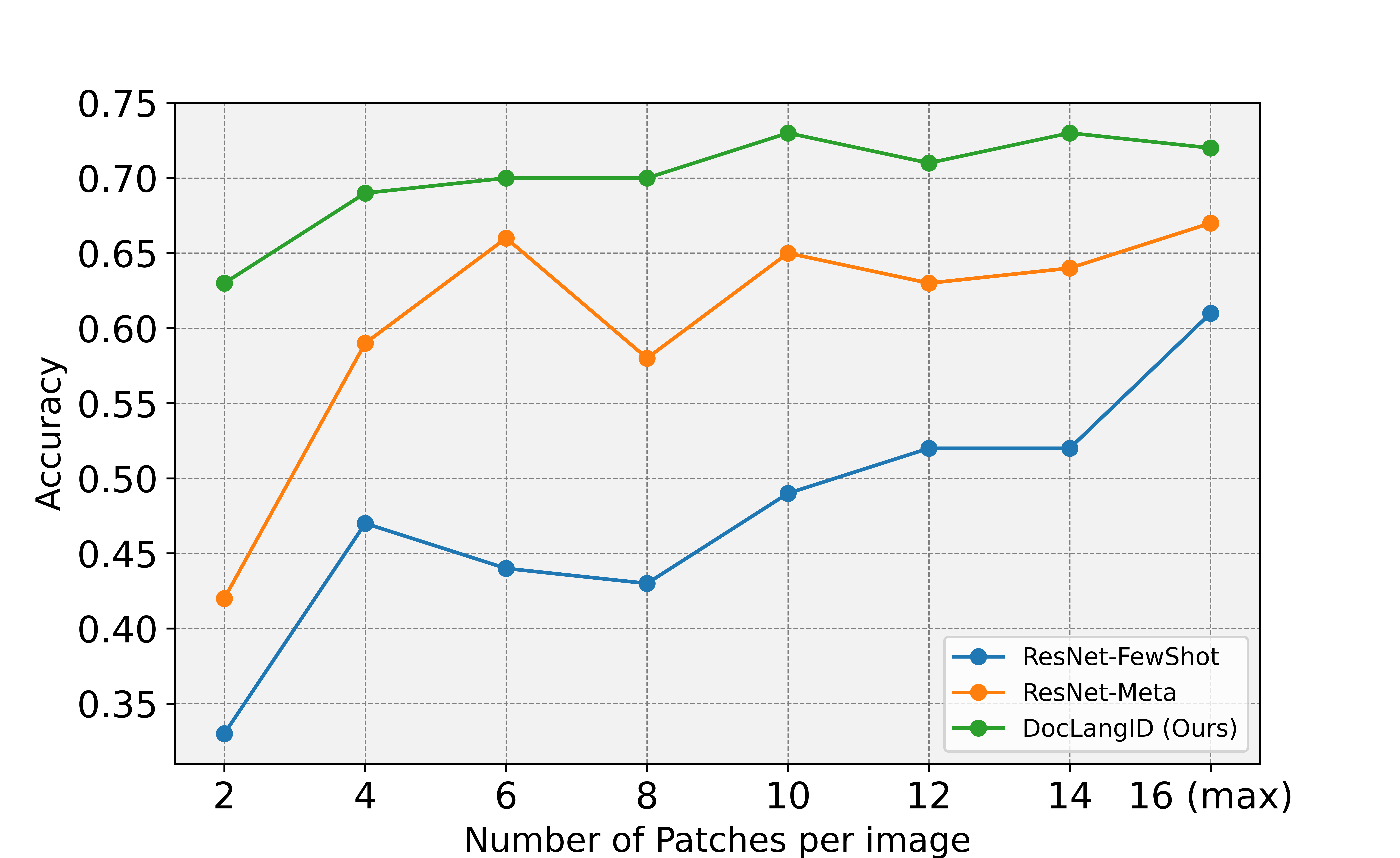}
    \caption{Influence of different numbers of patches on the recognition performance (accuracy scores using 50 few-shot samples).}
    \label{fig:patch_size_analysis}
\end{figure}

The core classification metric values for the \ac{WPI} target domain are presented in \autoref{tab:core_accuracies}. It can be seen that our combined approach outperforms the single few-shot learning variation \textit{ResNet-FewShot} by over 20 percentage points.
\begin{table}[h!]
\centering
\caption{Results for the \ac{WPI} evaluation dataset (50 few-shot samples)}
\label{tab:core_accuracies}
\begin{tabular}{@{}ccccc@{}}
\toprule
Method           & Accuracy & Precision & Recall & F1   \\ \midrule
ResNet-FewShot   & 0.52     & 0.49      & 0.54   & 0.51 \\
ResNet-Meta      & 0.64     & 0.68      & 0.69   & 0.68 \\
\textbf{DocLangID (Ours)} & \textbf{0.74}     & \textbf{0.77}      & \textbf{0.74}   & \textbf{0.74} \\ \bottomrule
\end{tabular}
\end{table} 
\autoref{fig:metrics_resnet} shows additional classification metric results for each target domain language. Again, we evaluated our \textit{DocLangID} approach using 50 few-shot samples per target domain language. Overall, our model yields high classification metric results, despite some minor differences between the language classes. Nevertheless, we observed that it was slightly more difficult for the model to distinguish between the German and Dutch language classes, leading to a lower recall value for Dutch. This may be related to higher linguistic similarities between German and Dutch. In terms of recognition accuracy, Tesseract generally achieved very good results ($> 90\%$), outperforming \textit{DocLangID} in this comparison in general. However, achieving this performance required us to input the correct set of languages beforehand, of course.
A critical factor for the performance of \textit{DocLangID} is the number of few-shot samples used for the two-stage training. \autoref{fig:the-samples} shows the effect of the different subsets of the 50 few-shot samples on the recognition performance. Generally, it can be seen that the recognition performance improves with an increasing number of few-shot samples and that \textit{DocLangID} outperforms the other variants for all configurations. Interestingly, even smaller amounts of samples per language already improve recognition performance substantially for all variants. 
Another important component of our approach is the number of patches considered during training. \autoref{fig:patch_size_analysis} shows the influence of different numbers of patches on the recognition performance. Our approach achieves almost 70\% recognition accuracy using only four patches. Although a higher number ($\ge 8$) of patches still led to higher accuracy, the performance improvement seems not to be very strong. In contrast, a smaller number of patches, such as two, strongly degraded the performance for \textit{ResNet-FewShot} and \textit{ResNet-Meta}. For the other experiments, we used the maximum patch number of 16, which yielded the best results for all methods. Especially for \textit{ResNet-FewShot}, this number yielded substantial recognition improvements. One interpretation could be that \textit{ResNet-FewShot}, in contrast to \textit{DocLangID} and \textit{ResNet-Meta}, is only trained on the few-shot samples and therefore, on a significantly smaller dataset. While this could lead to a more straightforward learning process on the target domain, this could also increase the risk of overfitting.
In contrast, we find that the numerous labeled examples of the IMPACT dataset positively influence the adaptability of the model, despite the domain differences that come from different image structures and language classes. Also, note that the maximum number of patches strongly depends on the size of one patch, and the values in our work represent a design choice. Thus, small amounts of patches with larger patch sizes could lead to similar observations.
Moreover, we showcased our approach to the domain shift between two domains only, which may be seen as a low number for domain adaptation. However, we considered two fundamental domain characteristics between our source and target domain: language and document type. Not only are the sets of languages (except for Dutch) for both domains disjoint, but the document types are also fairly different. While our \ac{WPI} target domain mostly covers images of auction catalogs, often only showing large pictures of the auction objects and comparatively few and small texts, our IMPACT \cite{impact_paper} source domain entails text-rich images of diverse documents, such as articles, newspapers, and pamphlets. In this context, we found that our source domain itself covers a sufficient amount of diverse characteristics that can be used to demonstrate the effectiveness of our approach to adapt to the challenging \ac{WPI} domain.
\subsection{Recognition speed}
Apart from the correctness of the prediction, another important qualitative aspect is the time needed to compute the prediction. In comparing \textit{DocLangID} with our baseline using Tesseract, we evaluated the average processing times at inference, shown in \autoref{tab:the_times_averages}. The values are calculated by an average of 100 raw images, where each language is equally represented. The table differentiates between the processing time needed for classification and the preprocessing steps described in subsection \autoref{datasets}.

The Tesseract approach requires almost ten times the amount of time for IMPACT. This trend becomes even more apparent for the \ac{WPI} dataset and when we deduct the preprocessing time. In this scenario, our \textit{DocLangID} model can compute the language classification for an image in just a few milliseconds, whereas Tesseract still needs more than one second. One interpretation for this large difference could be that, although the average file size for images from both datasets is similar, many of the IMPACT images contain more intricate details, text, and patterns as compared to the \ac{WPI} images. This, in turn, could lead to a higher image complexity and hence, higher computational costs to analyze a single image. Therefore, utilizing a more efficient, deep-learning-based solution like \textit{DocLangID} can be beneficial to overcome this potential \ac{OCR} bottleneck while still providing reasonably good language information. This information can then be used to select a language-specific \ac{OCR} engine, effectively reducing the number of time-intensive \ac{OCR} runs. We refer to the appendix for example images of both datasets and all languages considered in this work.
\begin{table}[H]
\caption{Inference times in seconds averaged over all language classes of both datasets, with (w) and without (w/o) input preprocessing time.}
\centering
\begin{tabular}{@{}cccccc@{}}
        \toprule
        & \multicolumn{2}{c}{\textbf{w} Preprocessing Time} &  & \multicolumn{2}{c}{\textbf{w/o} Preprocessing Time} \\ 
Dataset & Tesseract          & DocLangID         &  & Tesseract           & DocLangID           \\ \midrule
IMPACT  & 21.2               & \textbf{2.95}              &  & 19.9                & \textbf{1.18}                \\
WPI     & 7.5                & \textbf{0.63}              &  & 6.5                 & \textbf{0.003}               \\ \bottomrule
\end{tabular}
\label{tab:the_times_averages}
\end{table}
\section{Conclusions and Future Work}
In this work, we have proposed and extensively evaluated \textit{DocLangID}, a two-stage, deep-learning approach combining meta-learning and few-shot learning for language identification of historical documents. We have developed baselines for this task by considering variations of our main \textit{DocLangID} approach that do not use the abundant labeled examples of the source domain. Our evaluation of the recognition quality has shown that our combined approach yields the highest prediction accuracy on our \ac{WPI} target domain. We also compared the inference speed of \textit{DocLangID} with language identification performed by Tesseract and the Python library fasttext-langdetect. For both IMPACT and \ac{WPI} datasets, \textit{DocLangID} achieves an inference time which is more than ten times faster than Tesseract. 

Possible future works include improving the patch extraction procedure using text localization techniques and incorporating unsupervised learning techniques to learn even better image representations from the large amounts of unlabeled data.

\bibliographystyle{ACM-Reference-Format}

\bibliography{references}

\appendix
\include{appendix}

\end{document}